\documentclass[a4paper]{svproc}
\usepackage{cite} 
\usepackage{type1cm}        
\usepackage{makeidx}         
\usepackage{graphicx}        
\usepackage{multicol}        
\usepackage[bottom]{footmisc}

\usepackage{enumitem}
\usepackage{newtxtext}       %
\usepackage[varvw]{newtxmath}       

\usepackage{subfigure}
\usepackage{url}
\usepackage{wrapfig}

\usepackage{array}
\usepackage{graphicx}
\usepackage{xcolor}
\usepackage{hyperref} 

\makeatletter
\renewcommand\section{\@startsection{section}{1}{\z@}%
  {6pt plus 2pt minus 2pt}%
  {6pt plus 2pt minus 2pt}%
  {\normalfont\large\bfseries}}

\renewcommand\subsection{\@startsection{subsection}{2}{\z@}%
  {5pt plus 2pt minus 2pt}%
  {5pt plus 2pt minus 2pt}%
  {\normalfont\normalsize\bfseries}}

\renewcommand\subsubsection{\@startsection{subsubsection}{3}{\z@}%
  {5pt plus 2pt minus 2pt}%
  {5pt plus 2pt minus 2pt}%
  {\normalfont\normalsize\itshape}}
\makeatother

\usepackage{eso-pic}
\newcommand\AtPageUpperMyright[1]{\AtPageUpperLeft{%
 \put(\LenToUnit{0.5\paperwidth},\LenToUnit{-1.5cm}){%
     \parbox{0.5\textwidth}{\raggedright\fontsize{11}{11}\selectfont #1}}}}
\newcommand{\conf}[1]{\AddToShipoutPictureBG*{\AtPageUpperMyright{#1}}}

\makeindex             

\conf{\hspace*{-5cm}\textcolor{gray}{This paper has been accepted for publication at \textcolor{blue}{\href{https://iser2025.org/}{(ISER 2025)}}. }}

\begin{document}
\mainmatter
\title{MiniUGV$_2$: A Compact UAV-Deployable Tracked Ground Vehicle with Manipulation Capabilities 
\vspace{-10pt}}
\titlerunning{UAV-Deployable miniUG$V_2$}
\author{{Durgakant Pushp$^{1}$ \and Swapnil Kalhapure$^{2}$ \and Shaekh Mohammad Shithil$^{1}$ \and Lantao Liu$^{1}$}
\vspace{-5pt}
}

\authorrunning{D. Pushp et al.}

\institute{
}

\maketitle
\begin{abstract}
Exploring and inspecting \emph{Hidden Spaces}, defined as environments whose entrances are accessible only to aerial robots but remain unexplored due to geometric constraints, limited flight time, and communication loss, remains a major challenge. We present miniUGV$_2$, a compact UAV-deployable tracked ground vehicle that extends UAV capabilities into confined environments. The system introduces dual articulated arms, integrated LiDAR and depth sensing, and modular electronics for enhanced autonomy. A novel tether module with an electro-permanent magnetic head enables safe deployment, retrieval, and optional detachment, thereby overcoming prior entanglement issues. Experiments demonstrate robust terrain navigation, self-righting, and manipulation of objects up to 3.5 kg, validating miniUGV$_2$ as a versatile platform for hybrid aerial-ground robotics.
\footnote{Luddy School of Informatics, Computing, and Engineering at Indiana University, Bloomington, IN 47408, USA. 
Email: {\tt\small \{dpushp, sshithil, lantao\}@iu.edu}. 
$^{2}$S. Kalhapure is an Independent Researcher at the time of work. 
Email: {\tt\small swapnil.kalhapure@outlook.com}. 
The authors appreciate for supports from ERDC with Grant \#W912HZ2320003.}
\end{abstract}

\section{Introduction}
\label{sec:intro}

The exploration and inspection of complex and confined environments such as post-earthquake collapsed structures~\cite{hu2019detecting, nedjati2016post, polka2017use}, rock-blasting sites~\cite{roy2005rock, BAJPAYEE200447}, and underground culverts~\cite{mitchell2005risk, asadi2020integrated} remain significant challenges for autonomous robotics. Ground robots excel at traversing rugged terrain and performing manipulation tasks, but they often lack the mobility to reach elevated or obstructed entry points of the area to be inspected. Conversely, aerial robots can rapidly access such locations but face constraints in payload capacity and flight endurance.
A key challenge arises in the exploration or inspection of what we refer to as {\em Hidden Spaces} where the entrance is accessible only by flying robots, but the interior cannot be fully explored by them due to geometric constraints (e.g., narrow entryways), limited flight time, communication degradation, and unknown internal traversability. These scenarios demand a hybrid aerial-ground system that combines the accessibility of UAVs with the exploration and manipulation capabilities of ground robots.

While existing UAV-UGV collaboration frameworks often treat the ground robot as the primary system, using the drone merely as a ``flying sensor," this approach fails to fully address the challenges posed by Hidden Spaces. Our prior work~\cite{pushp2022uav} introduced a counterintuitive system that inverted this hierarchy: the UAV served as the primary agent, deploying a miniature ground robot to perform inspection and manipulation tasks in confined environments. Despite its promise, this system had several limitations: the miniUGV lacked sufficient size and sensor capacity to handle more complex inspection and manipulation tasks; it was permanently tethered to the drone, which restricted its autonomy; and the simple motor-pulley system for the tether mechanism frequently suffered from entanglement issues with the motor shaft, especially during UAV roll and pitch maneuvers while winding or unwinding the tether.

To address these limitations, we present {\em miniUGV$_2$}: a compact, UAV-deployable tracked ground vehicle equipped with advanced sensing and manipulation capabilities. This new prototype features two independently controlled, fully articulated arms capable of 360-degree operation, a suction-based gripper for versatile object handling, addressable LED headlights for low-light visibility, and integrated 2D LiDAR and depth sensors for enhanced autonomy. Additionally, a novel tether module with an electro-permanent magnetic head enables the UAV to deploy and retrieve the miniUGV$_2$ on demand, supporting both \textit{attached mode} where the UAV and miniUGV remain connected via the tether and \textit{detached mode} where the tether is disengaged and both platforms operate independently. 
\begin{figure}[t!]
    \centering
    \includegraphics[height=1.7in, width=4.6in]{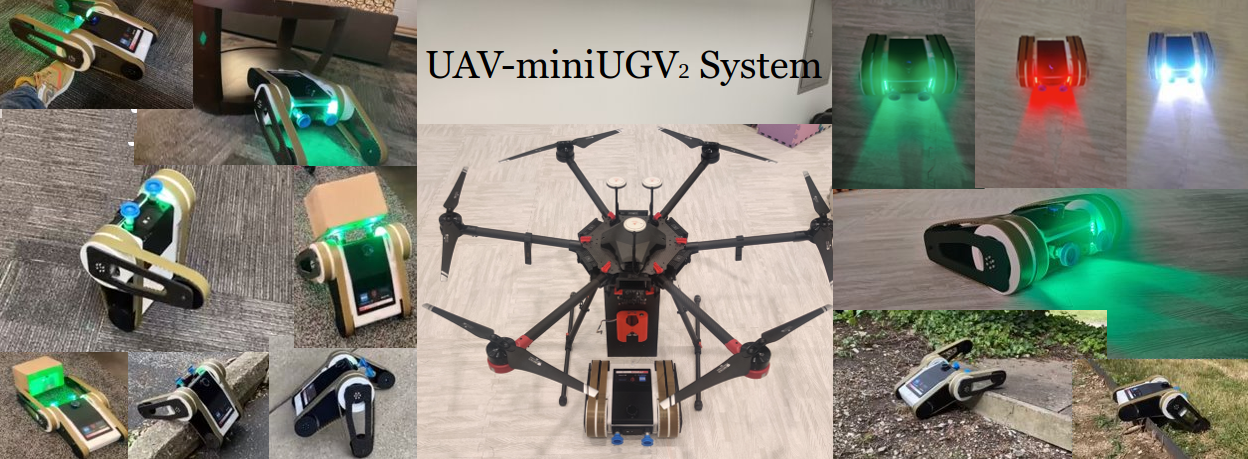}
    \caption{Illustration of the UAV-miniUGV system: Left and right images depict the designed miniUG$V_2$, an improved version of the miniUGV~\cite{pushp2022uav}. Center image shows the primary and secondary robots together with a tether module.
    }
    \label{fig:enter-label}\vspace{-12pt}
\end{figure} 
\\
\\
The key contributions of this work include:
\begin{enumerate}[leftmargin=*]
    \item \textbf{A Compact miniUG$\mathbf{V_2}$:} 
    We design and prototype a tracked ground vehicle that maintains UAV deployability while incorporating advanced manipulation and sensing capabilities. We also provide all necessary resources for replication. 

    \item \textbf{Tether Module with Active Tether Head:} 
    We have developed a novel tether module that uses a single actuator to control both the tether rotor and tether guider. The tether head features an electro-permanent magnet (EPM), which can be activated or deactivated via WiFi, enabling the drone to attach or detach the miniUG$V_2$ on demand.

    \item \textbf{Software Integration Framework:} 
    We provide a software architecture that integrates the drone, tether-pulley mechanism, and magnetic tether head within a ROS framework, complemented by a MuJoCo-based drone–tether–module simulator. This combination enables researchers to test core functionalities, prototype control and learning algorithms, and establish a foundation for bridging the gap between conceptual design and real-world aerial–ground deployments.

\end{enumerate}

By merging a more capable miniUG$V_2$ with an innovative magnetic tether head and a high-fidelity simulation platform, our proposed system significantly enhances the operational capabilities of UAV-ground robot teams.
Supplementary materials, including software, and simulation environments, are publicly available at: \url{https://github.com/Dpushp/drone-deployable-miniUGV2}.


\begin{figure*}[t!] 
    \centering
    \includegraphics[width=4.6 in]{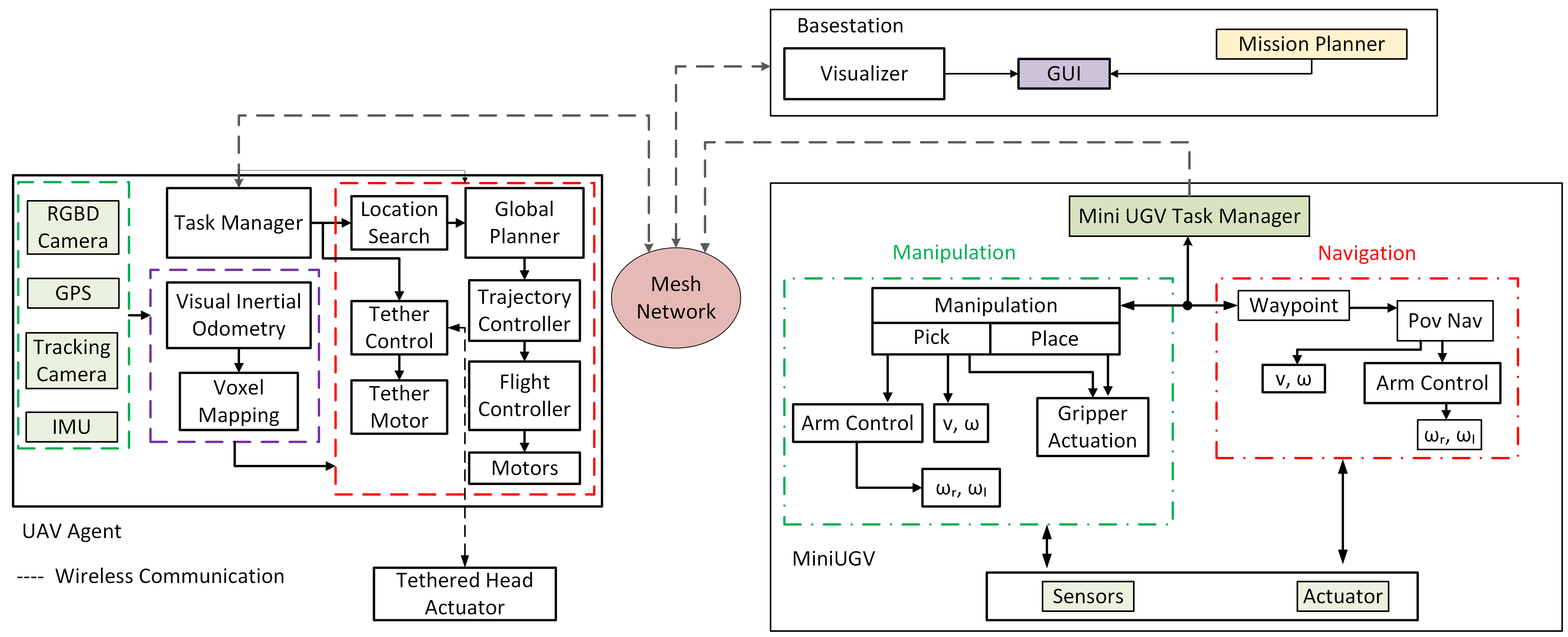}
    \caption{Proposed system architecture of the UAV-miniUGV platform for collaborative exploration. The UAV is equipped all the entire sensor suite to create a 3D mapping of the environment and find potential deployable zones for the miniUGV. The UAV then autonomously deploy and retrieve the robot using magnetic tether head. The miniUGV has articulated arms for manipulation and navigation in rough terrains. Both robots share their trajectory information via WiFi network and the high level mission command is sent from the Base station.}
    \label{fig:system_overview} \vspace{-12 pt}
\end{figure*}

\section{System Overview}
\label{sec:system_overview}

The proposed system comprises three core components: UAV, a tether module with an active tether head, and a miniature ground vehicle. Building upon our previous work~\cite{pushp2022uav}, we leverage and extend the existing autonomy stack for the UAV and tether module, adapting it to meet the new requirements imposed by the deployment and retrieval of the enhanced miniUGV$_2$.

As illustrated in Fig.~\ref{fig:system_overview}, the UAV initiates the mission by scanning the deployment area and identifying a suitable surface near the entrance of the {\em Hidden Space} that is navigable for the miniUGV$_2$. Once an appropriate deployment site is located, the UAV engages the tether module, releasing the tether until the miniUGV$_2$ reaches the ground or is within a safe proximity to it. The design of the miniUGV$_2$ ensures that it can operate effectively in any orientation, with the ability to self-correction if flipped during deployment. Consequently, the decision to detach the tether can be made at any point when the miniUGV$_2$ is sufficiently close to the ground to guarantee a safe landing.
For this purpose, we enforce a hard constraint on the minimum distance between the miniUGV$_2$ and the ground plane prior to detachment, ensuring deployment safety. The UAV utilizes an RGB-D camera to map the environment, locate the deployment zone, monitor the miniUGV$_2$ during descent and touchdown, and estimate its precise landing position. Following deployment, the miniUGV$_2$ transitions to autonomous operation, employing an image-based planning and control method to navigate within the Hidden Space.

Upon completion of its ground operation, the miniUGV$_2$ exits the Hidden Space and requests retrieval by the UAV. The UAV, guided by a visual servoing algorithm, aligns itself above the tether attachment point and releases the tether. The electro-permanent magnet within the tether head enables automatic attachment to the miniUGV$_2$ upon contact. Finally, the UAV retracts the miniUGV$_2$ and transports it back to the base. The detailed autonomy and control strategies of the system are elaborated in the following subsections.

\subsection{UAV Autonomy Stack}
\label{subsec:uav_autonomy_stack}

The UAV serves as the primary agent responsible for deploying and retrieving the miniUGV$_2$. Its autonomy stack is adapted from our previous work~\cite{pushp2022uav} and is now extended to accommodate dynamic Hidden Space deployment scenarios. The autonomy stack has 3 core components. 1. Sensor fusion-based localization, 2. Planning and control, and 3) MiniUGV deployment and retrieval.   

The UAV is equipped with GPS, RGB-D, and a Tracking camera for sensor fusion-based localization. We use Visual-Inertial Navigation System (VINS) fusion for the localization~\cite{qin2019a} of the UAV. The Voxel Mapping module creates a dense 3D map of the environment. The global planner takes the high-level mission plan as input and generates a feasible path toward the mission goal. The global path is executed by a B-spline trajectory generation module which creates dynamically feasible paths for the drone. A PID-based trajectory tracking controller tracks the B-spline trajectory. The deployment location search module finds a suitable location based on terrain elevation and image segmentation. 

Once a suitable deployment area is identified, the UAV engages its tether module, carefully lowering the miniUGV$_2$ while maintaining flight stability. A PID-based velocity controller regulates the rate of tether release, ensuring controlled descent. During deployment, the UAV continues to monitor the environment and the miniUGV$_2$ using real-time perception data, facilitating safe and precise touchdown. Following the completion of the miniUGV$_2$’s ground operation, the UAV transitions to retrieval mode. It re-aligns itself above the tether attachment point using visual servoing and releases the tether to re-engage the miniUGV$_2$. The electro-permanent magnet within the tether head ensures reliable reattachment. Once secure, the UAV retracts the tether, lifting the miniUGV$_2$ and completing the mission. 

\subsection{miniUGV$_2$ Autonomy}
\label{subsec:ugv_autonomy}

The miniUGV$_2$ operates autonomously within the Hidden Space once deployed. It is equipped with an onboard computational platform capable of executing image-based planning and control algorithms, enabling it to navigate through complex and confined environments. The autonomy stack leverages RGB-D and LiDAR sensors to generate local environment maps and identify traversable regions.
Path planning is achieved using a variant of the POVNav framework~\cite{pushp2023povnav, pushp2025navigating}, which segments the environment into navigable and non-navigable areas based on surface normal estimation. The miniUGV$_2$ then selects safe trajectories that maximize area coverage while avoiding obstacles. A low-level controller translates the planned path into motor commands using visual servoing.

To enhance maneuverability over uneven terrain, an {\em Arm Controller} module estimates the height of the tallest peak in the immediate environment and calculates the local slope relative to the miniUGV$_2$ body. It then adjusts the articulated arm angles accordingly to facilitate smoother traversal. While a more sophisticated control approach, explicitly considering terrain traversability~\cite{terrain_traversability, chen2022cali, chen2023pseudo, khan2025afrda}, wheel-ground interaction and traction forces~\cite{zou2018dynamic} would be required for full system deployment, this prototyping implementation uses a simpler control strategy that serves the immediate purpose.

During operation, the miniUGV$_2$ periodically assesses its relative position to the Hidden Space exit point to ensure it can return for retrieval. Once the exploration task is completed, or a predefined time or energy budget is reached, the miniUGV$_2$ navigates back to the exit and signals the UAV for retrieval.

\subsection{Tether Module Control}
\label{subsec:tether_module_control}

The tether module is responsible for the controlled deployment and retraction of the miniUGV$_2$. It consists of a single actuator that simultaneously manages both the tether rotor and the tether guider, simplifying mechanical complexity while ensuring reliable operation. A PID-based velocity controller modulates the actuator speed, maintaining a steady deployment rate that adapts to dynamic UAV motion. 

We propose a position tracking system to track the tethered module for precise deployment and retrieval operations. It fuses depth point cloud and tether rotor encoder feedback to estimate the tether head position relative to the drone as shown in Fig. 3.
It performs real-time tracking using Euclidean clustering DBSCAN algorithm~\cite{deng2020dbscan}.  It identifies the miniUGV by selecting the most centered and closest cluster within area of interest and estimating the centroid to find the position.  The tethered encoder provides accurate feedback of the vertical position in case of vision occlusion, which makes the system robust in unstructured environments.  
\begin{wrapfigure}{r}{0.6\textwidth}
    \vspace{-10pt}
    \centering
    \includegraphics[width=\linewidth]{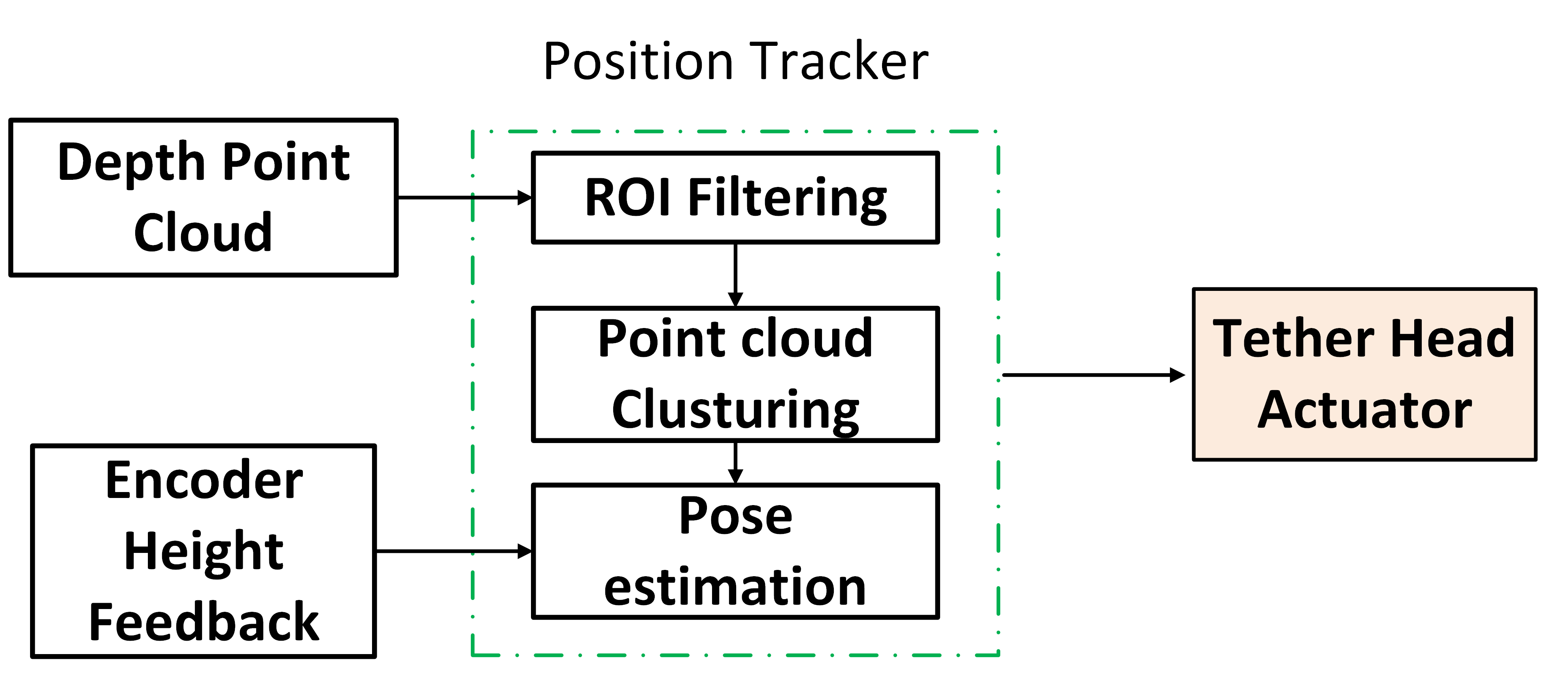} 
    \caption{\small Overview of the tether head position tracking.}
    \label{fig:tether module}
    \vspace{-20pt}
\end{wrapfigure}

The tether head features an electro-permanent magnet that can be activated or deactivated via WiFi commands from the UAV’s onboard computer. For retrieval, the UAV aligns itself with the miniUGV$_2$’s attachment point using visual servoing and releases the tether to allow the magnetic head to reattach autonomously. It is important to note that this reattachment procedure requires precise positioning of the head, and thus has not yet been validated in real outdoor conditions due to its complexity; it is therefore considered a separate problem for future work. Once the attachment is confirmed, the tether module retracts the miniUGV$_2$ to complete the operation. 

\section{Hardware System Realization}
\label{sec:hardware_realization}

This section details the hardware implementation of the proposed system, with the aim of ensuring reproducibility and facilitating future research efforts. The primary robot is selected from commercially available UAV platforms that satisfy two key requirements: a minimum payload capacity of at least 6 kg and an attachment area exceeding $350\text{mm} \times 350\text{mm}$. These specifications ensure compatibility with widely used UAV models such as the DJI Matrice 600~\cite{dji_matrice600pro}, which provide the necessary capacity and physical space to carry the secondary robot as a payload. Alternatively, any custom-built UAV meeting these requirements can serve as the primary robot. The remainder of this section describes the prototyping details of the miniature ground vehicle and the tether module, both of which can be readily integrated with any compliant UAV platform to complete the system.

\subsection{Secondary Robot (miniUG$V_2$)}
\label{subsec:hardware_design}

\begin{figure*}[t!] \vspace{-5pt}
{
    \centering
    \subfigure[miniUG$V_2$]
        {\label{fig:top} \includegraphics[height=1.5in, width=2.8in]{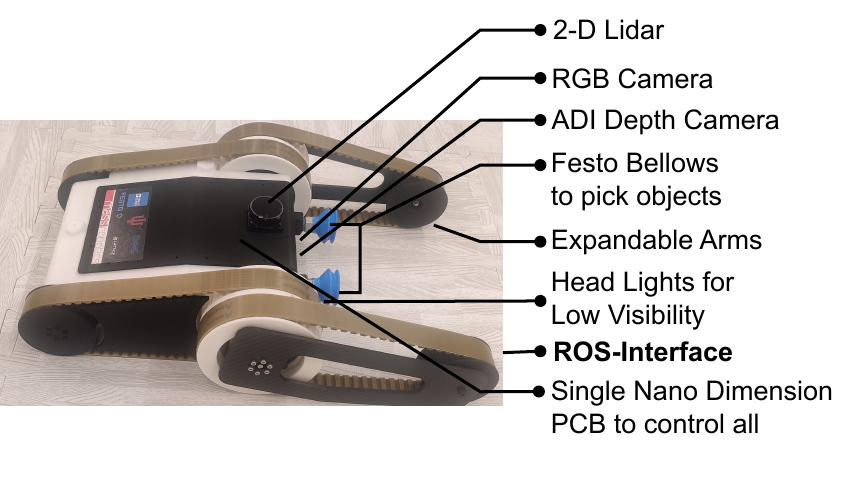}}
    \subfigure[Designed Circuit Board]
        {\label{fig:iso} \includegraphics[height=1.4in, width=1.3in]{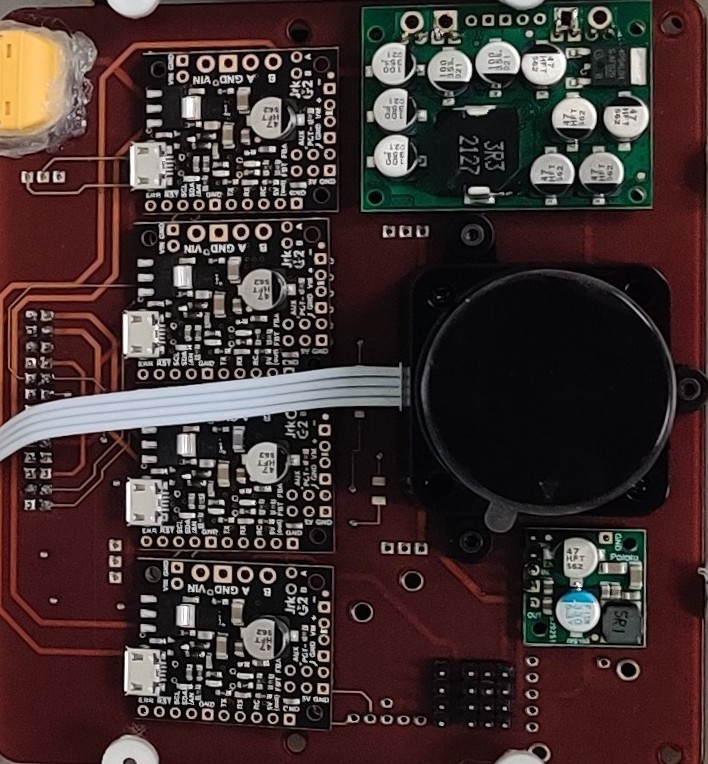}}
    \vspace{-8pt}
    \caption{(a) Assembled miniUG$V_2$, showcasing its features, highlighting its compact size with integrated sensors and gripper mechanism. (b) Shows the designed circuit board with all the electronic components required to operate the miniUG$V_2$.\vspace{-12 pt}}
    \label{fig:decision_space}
}
\end{figure*}

\noindent\textbf{\textit{Design.}}
The design choice of miniUG$V_2$ incorporates a modular approach to meet size constraints and accommodate necessary sensors for autonomy. The compact chassis and modular bottom lid facilitate the integration of various payloads, such as grippers, depth cameras, RGB cameras, and localization cameras. The modular design allows for easy replacement of the bottom lid with different payloads, without requiring a complete redesign of the robot. Additionally, the decision to include extended articulated arms enables miniUG$V_2$ to navigate rugged surfaces while still adhering to the specified size constraint. To optimize the design of miniUG$V_2$, a front-free wheel configuration was implemented to accommodate the motor for controlling the articulated arm. This strategic placement of the motor within the front wheel creates additional space for the bottom lid, allowing for easy integration of various payloads. This design feature enhances the overall functionality and compactness of miniUG$V_2$, making it well-suited for its intended tasks.

\noindent\textbf{\textit{Integrated Circuit Board Design.}}
The miniUG$V_2$ robotic platform incorporates a custom-designed 6-layer Printed Circuit Board (PCB) that adheres to industry standards and optimizes functionality. 
It includes four motor drivers, three voltage regulators for reliable performance, and key features such as a 2D Lidar, LED interface, SD card slot, open 5V source, microcontroller interface, and I$^2$C communication port.

\noindent\textbf{\textit{Software Development: From Low-level control to ROS API.}}
The software development process for the miniUG$V_2$ encompasses low-level control and the development of a Robot Operating System (ROS) API. At the low-level control, the firmware was developed to interface with the microcontroller, enabling direct control over hardware components and implementing motor control algorithms, sensor data acquisition, and communication protocols. The firmware was optimized for efficient resource utilization and real-time responsiveness.
The development process then transitioned to integrating the miniUG$V_2$ system with the ROS framework. Developing a ROS API made the platform compatible with various ROS-enabled tools and libraries, allowing seamless integration with other robotic systems and enabling higher-level functionalities. The miniUG$V_2$ platform features left and right sides with 7 addressable LEDs each, and these LEDs can be controlled using the defined topic. This functionality enhances the robot's visibility in dark environments and provides status indicators for sensors, pump actuation, and arm motion. Additionally, other standard topics specific to the miniUG$V_2$ platform were included in the ROS API development.

\subsection{Deployment System}
\begin{figure*}[t!] \vspace{-5pt}
{
    \centering
    \subfigure[Tether Module]
        {\label{fig:tether-module} \includegraphics[height=1.6in, width=3.5in]{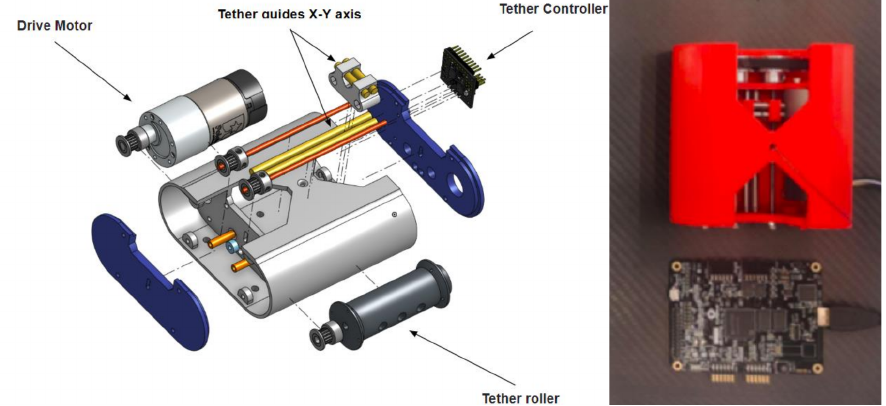}}
    \subfigure[Tether Head]
        {\label{fig:tether-head} \includegraphics[height=1.6in, width=1in]{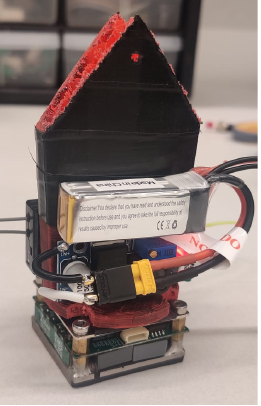}}
    \vspace{-8pt}
    \caption{(a) Shows the the tether module design on the left side and the actual prototype (3D printed using the red color PLA material) with the tether control onboard computer on the right side.(b) Shows the magnetic tether head. It is a fully autonomous ROS-enabled module incorporating an ESP32 microcontroller, a LiPo battery for power supply, and a voltage regulator that distributes power to both 5V and 3.3V rails.\vspace{-12 pt}}
    \label{fig:deployment-system}
}
\end{figure*}

The deployment mechanism, as shown in Fig.~\ref{fig:deployment-system}, consists of a compact tether module and a tether head. The tether module uses a single actuator to simultaneously control both the tether rotor and the tether guider, simplifying the mechanical design and reducing the overall payload. The tether head integrates an electro-permanent magnet, which can be remotely activated or deactivated by the UAV via WiFi over ROS. This enables precise and reliable attachment or detachment of the miniUG$V_2$, allowing the ground robot to operate independently once deployed.  

Retrieval of the miniUG$V_2$ requires reattachment via the magnetic tether, a task demanding more advanced control strategies beyond the scope of this paper. To facilitate further research, we provide a realistic MuJoCo simulation environment for addressing these challenges, available at the shared GitHub repository (see Section~\ref{sec:intro}).

\begin{table}[h!] \vspace{-0.2cm}
\centering
\caption{Comparison of miniUG$V_2$ with Existing Mobile Platforms}
\begin{tabular}{|c|c|c|c|c|c|c|c|}
\hline
\textbf{No.} & \textbf{Robot} 
& \parbox{1.5cm}{\centering \textbf{Dimension\\L/W/H(mm)}} 
& \parbox{1cm}{\centering \textbf{Weight\\(kg)}} 
& \parbox{1.4cm}{\centering \textbf{Operation \\Time(hrs)}} 
& \parbox{2.cm}{\centering \textbf{Sensors, actuators \& \\Computer}} 
& \textbf{\centering Gripper} 
& \parbox{1cm}{\centering \textbf{ Active \\Flippers }}\\
\hline

1 & \parbox{2.cm}{\centering EyeDrive~\cite{eyedrive}\\
\includegraphics[width=2.cm, height=0.85cm]{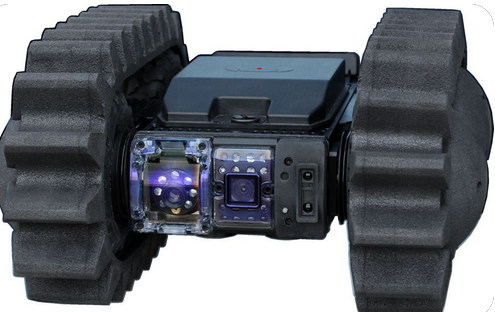}}
& 280/250/100
& 2.3
& 3.0
& \parbox{2.cm}{\centering ~\\1. 5 Cameras, \\2. IMU, \\3. Microphone \\Remote-\\Controlled\\~}
& NA
& No \\
\hline

2 & \parbox{2.cm}{\centering miniUG$V_1$~\cite{pushp2022uav}\\
\includegraphics[width=2.cm, height=0.85cm]{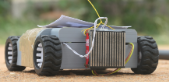}}
& 130/120/55
& $<$ 0.25
& 0.3
& \parbox{2.cm}{\centering ~\\1. RGB Camera, \\2. IMU, \\Autonomous-\\RPi Zero\\~}
& Magnetic 
& No \\
\hline

\textbf{3} & \parbox{2.cm}{\centering \textbf{ miniUG$\mathbf{V_2}$\\(Proposed)}\\
\includegraphics[width=2.cm, height=2.6cm]{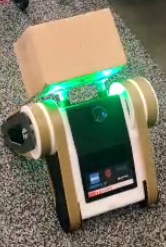}}
& \parbox{1.5cm}{\centering \textbf{330/330/100 with \\flippers\\ 490/330/100}} 
& $\mathbf{3.68}$
& $\mathbf{\approx 4.00}$
& \parbox{2.cm}{\centering ~\\1. RGB Camera, \\2. ToF Camera, \\3. IMU, \\4. 2D Lidar, \\5. Head Lights, \\6. Air pumps, \\7. TinyPICO, \textbf{\\Autonomous: \\Nvidia Jetson\\ Orin NX}\\~}
& \textbf{Pneumatic}
& \parbox{1cm}{\centering \textbf{Yes\\ 2\\motors}}\\
\hline

4 & \parbox{2.cm}{\centering 510 PackBot~\cite{packbot-robot} \\ \includegraphics[width=1.8cm]{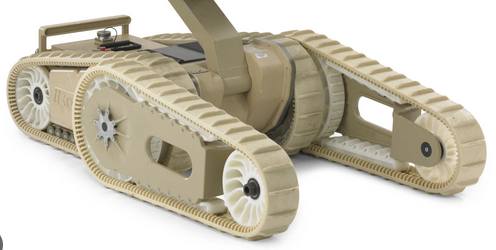}} 
& \parbox{1.5cm}{686/406/178 with flippers\\ 889/521/178} 
& $10.9$ 
& $\approx 4.00$
& All
& Any
& \parbox{1cm}{\centering Yes\\ 2\\motors}\\
\hline

5 & \parbox{2.cm}{\centering Dr. Robot~\cite{jaguar-mobile-platform} \\ \includegraphics[width=1.8cm]{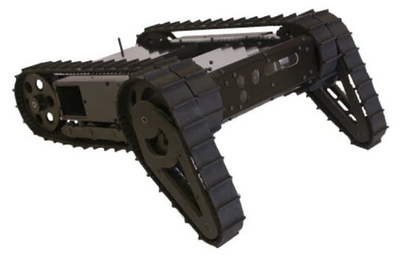}} 
& \parbox{1.5cm}{640/655/176 with flippers\\ 820/655/176} 
& $21.5$ 
& $4.00$
& All
& Any
& \parbox{1cm}{\centering Yes\\ 2\\motors}\\
\hline

\end{tabular}
\label{table:comparison}\vspace{-0.4cm}
\end{table}

\section{Experimental Results}
\label{sec:experimental_results}

This section presents a comparative study of the miniUGV$_2$ with existing mobile robotic platforms and provides experimental results that rigorously evaluate the functionality and performance of the proposed system. The experiments focus on two primary aspects: the navigation and manipulation capabilities of the miniUGV$_2$, and its successful deployment for hidden area exploration using the proposed tether module. 

\subsection{MiniUGV$_2$}
\textbf{Comparative Study: }We surveyed  several existing wheel-based robotic platforms and conducted a comparative analysis between them and our proposed miniUGV$_2$, focusing specifically on their suitability as secondary robots deployable by UAVs. The evaluation considered key parameters, including overall dimensions (to ensure compatibility with the UAV’s payload attachment area), weight (to comply with the UAV’s payload capacity), operational time (to determine practical usability in real-world scenarios such as culvert inspection), sensor and actuator configurations, and the presence of companion computers supporting autonomous operations.

As summarized in Table~\ref{table:comparison}, the EyeDrive and miniUGV$_1$ platforms meet the deployability criterion (i.e., payload less than 6 kg) but fall short in several critical areas, including operational time, sensor capabilities, actuation mechanisms, onborad computing, gripper functionality, and active flipper modules. Conversely, the PackBot and Dr. Robot platforms offer enhanced capabilities in terms of sensors and actuation but are unsuitable for UAV deployment due to their heavy weight and larger physical size.
By contrast, miniUGV$_2$ achieves a distinctive balance of portability and capability, establishing it as an ideal secondary robot for UAV-supported hidden space exploration.

\begin{figure}[t!] 
    \centering
    \includegraphics[width=4.6in]{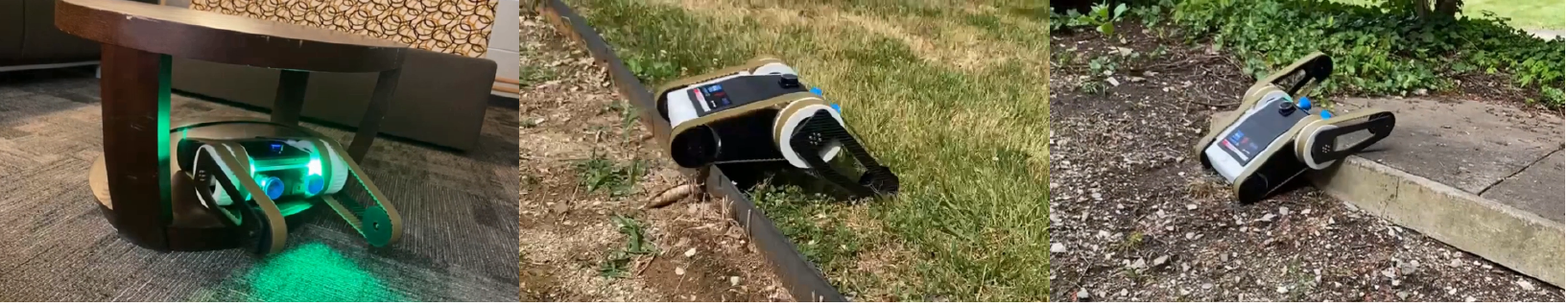}
    \caption{Illustration of the miniUGV$_2$ navigating three distinct scenarios. From left to right: the first image demonstrates its navigation capabilities within a cluttered indoor environment; the second image showcases the miniUGV$_2$ traversing a sharp border edge which is a particularly challenging task for a robot of this form factor; and the third image highlights its ability to navigate uneven terrain composed of gravel and concrete structures.
    \vspace{-12pt}
    }
    \label{fig:miniUGV_nav}
\end{figure}

\textbf{Navigation Tests:}  
For navigation testing, we evaluated the miniUGV$_2$ across a variety of indoor and outdoor environments, including rough and smooth surfaces, as well as uneven terrains featuring high and sharp obstacles, as shown in Fig.~\ref{fig:miniUGV_nav}. Note that we used manual drive to demonstrate these capabilities of the prototype robot. The miniUGV$_2$ demonstrated exceptional maneuverability during the navigation experiments, successfully traversing high obstacles, sharp border edges, and uneven surfaces. Its unique ability to operate in both upright and inverted configurations, combined with its capability to regain a default orientation using its dual arms, highlights its adaptability to challenging terrains. The robust design effectively protects all onboard sensors, ensuring uninterrupted functionality throughout navigation tasks.

\begin{wrapfigure}{r}{0.48\textwidth}
    \vspace{-20pt}
    \centering
    \includegraphics[width=\linewidth]{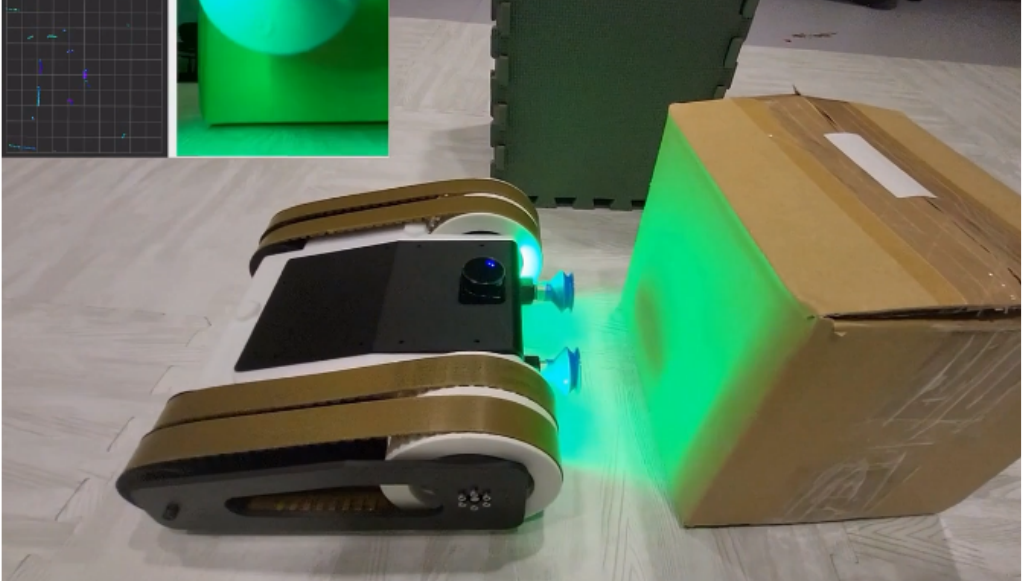}
    \caption{\small Illustrates the camera, LiDAR views and the position of the articulated arms before initiating a pick operation. } 
    \label{fig:miniUGV_man_test} \vspace{-15pt}
\end{wrapfigure}
\textbf{Manipulation Tests:}
Manipulation testing involved the use of boxes of varying sizes and weights ranging from 0.5 kg to 3.5 kg. 
The miniUGV$_2$ exhibited reliable object handling capabilities, successfully picking up and placing boxes of different sizes. These results validate the miniUGV$_2$ as a versatile and dependable secondary robot, suitable for various UAV-supported missions. During pick and transport operations, the miniUGV$_2$ is required to restrict its arm movement between 0 and 90 degrees to prevent interference with the suction grippers, particularly when handling objects larger than the clearance area between the arms, as shown in Fig.~\ref{fig:miniUGV_man_test}. In such cases, the miniUGV$_2$ maintains a default arm configuration to maximize the contact area and consequently the traction force, thereby enhancing its ability to manipulate and transport heavier objects effectively.

\subsection{Autonomous Deployment of MiniUGV$_2$}

\begin{figure}[t!] \vspace{-5pt}
    \centering
    \includegraphics[width=4.6in]{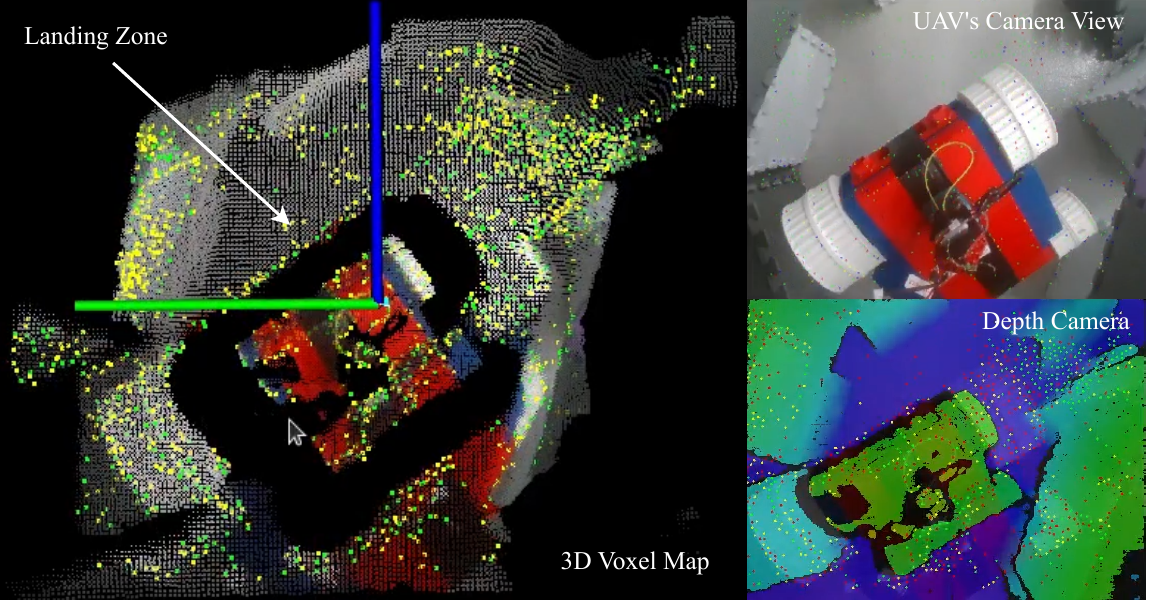}
    \caption{Illustration of the initial state of UAV deploying the miniUGV$_2$: This figure captures the snapshot just before the deployment process begins. The left panel shows the point cloud generated from the UAV’s downward-facing RGB-D camera. The coordinate frame at the center of this image represents the camera frame. Yellow points correspond to visual features extracted from the RGB image and projected into 3D space. These same visual features are highlighted in the other two images. The top-right panel displays the raw RGB camera view, while the bottom-right panel shows the colorized depth image, where blue points indicate areas farther from the camera and green points indicate areas closer to it.
    } \vspace{-12pt}
    \label{fig:landing_start}
\end{figure}
We conducted a deployment experiment in which the UAV autonomously flies over the target area and uses its onboard sensors to identify the optimal deployment point - defined as the closest navigable surface for the miniUGV$_2$ from the entry point of the Hidden Space targeted for inspection or exploration. The deployment surface is determined by fitting a plane to the point cloud generated by the UAV’s downward-facing depth camera. In scenarios involving uneven terrain, a surface-normal-based image segmentation approach can be employed to detect the deployment zone (see the blue region in the depth image of Fig. \ref{fig:landing_start}). The surface normals generated from the depth provides a good idea about the navigability as the surfaces with same normal vector is colored the same and then a color segmentaation and other image processing toolts can be utilized to get the deployment area. 
As illustrated in Fig. \ref{fig:landing_start}, the UAV performs visual-inertial localization and mapping prior to initiating the deployment, generating a 3D map of the environment.
Point cloud clustering algorithms, combined with feedback from the tether controller motor, enable real-time tracking of the deployment payload. It is important to note that mapping is performed only prior to the initiation of the deployment process and it is resusmes after touchdown. Mapping is intentionally halted to prevent the inclusion of the moving miniUGV$_2$ in the map, thereby preserving a clean, static reference map for accurate landing verification and reducing computational overhead on the UAV’s onboard computer. 

\begin{figure}[t!] 
    \centering
    \includegraphics[width=4.6in]{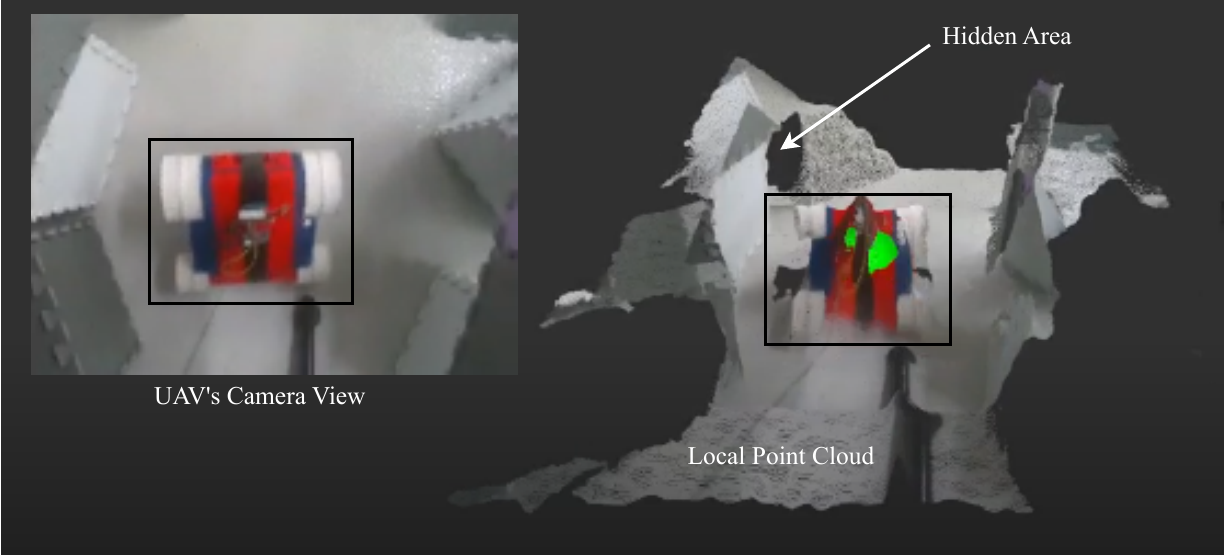}
    \caption{This Figure illustrates the scenario after landing when the 3D mapping process is resumed. Note that the map update is paused during the deployment. The posistion of miniUGv$_2$ is determined by the point cloud processing and shown with green sphare. The colored point cloud is shown to verifiy that the position estimation is accurate and it is inside the point cloud cluster of the miniUGV$_2$. The opening of the hidden area is also clearly visible in the 3D map. The black square shown the miniUGV$_2$ in both the RGB image and the 3D point cloud map for reference.
    \vspace{-12pt}
    }
    \label{fig:landing_end}
\end{figure}

The point cloud after landing is shown in Fig.~\ref{fig:landing_end}. The landing process can be verified in several ways. For instance, if the distance between the tether head position and the miniUGV$_2$ exceeds a minimum threshold value (accounting for position estimation errors) when the UAV initiates the tether retraction and this distance continues to increase while the miniUGV$_2$ remains stationary, the deployment is considered successful. Otherwise, if this criterion is not met, the deployment is considered as failed, and the UAV reattempts the deployment. In our experiments, we primarily used the distance of the miniUGV$_2$ from the ground plane to identify a successful landing, as depicted in Fig.~\ref{fig:landing_end}, where the green sphere representing the miniUGV$_2$ position indicates proximity to the ground. This distance is tunable parameter for the field experiments. 
Additionally, we conducted more experiments to validate the tether module’s functionality under laboratory conditions to ensure reliable deployment and retrieval of the miniUGV$_2$. 
Detailed experimental setups and results are available at \url{https://github.com/Dpushp/drone-deployable-miniUGV2}.

\section{Conclusion}
\label{sec:5}

In this work, we presented miniUGV$_2$, an advanced UAV-deployable tracked ground vehicle designed to tackle the unique challenges of exploring and inspecting {\em Hidden Spaces}. Building upon our prior work~\cite{pushp2022uav}, we introduced substantial improvements to both the ground robot and its deployment system, enabling more robust and autonomous operations in complex settings.
Our key contributions include the design of miniUGV$_2$ featuring dual fully articulated arms, advanced sensing capabilities, and modular electronics, which collectively enhance the robot’s maneuverability and manipulation proficiency. We also developed a novel tether module with an electro-permanent magnetic head, allowing the UAV to deploy and retrieve the miniUGV$_2$ on demand while supporting both attached and detached operational modes. This feature eliminates the tether entanglement issues of our previous system while retaining the option to operate in a permanently attached configuration to exploit additional thrust from the UAV when needed.
Although retrieving the miniature robot using the tethered system in real-world outdoor settings remains challenging, we have developed a high-fidelity MuJoCo-based simulation of the UAV–tether deployment system. This simulation provides a safe and reproducible testbed, which we plan to leverage for the development and validation of advanced retrieval algorithms in future work.
Experimental validation demonstrated the miniUGV$_2$'s capabilities across diverse scenarios, including navigating rough terrains, self-righting after flipping, and manipulating objects weighing up to 3.5 kg. While the arm was manually controlled during field experiments since the simple control logic outlined in this work is insufficient for highly challenging environments; the results nevertheless demonstrate the system’s core capabilities and establish a solid foundation for future research on fully autonomous operation. Furthermore, both laboratory and real-world tests verified the reliable performance of the tether system and its integration with the UAV for deployment and retrieval.

We plan to extend this work by integrating advanced control strategies for miniUGV$_2$ that explicitly consider terrain traversability, traction forces, and real-time environmental changes. Furthermore, future studies will explore the development of autonomous decision-making frameworks inspired by bounded rationality principles~\cite{pushp2024context, xu2022decision, pushp2023coordination}, with the aim of enhancing the system's adaptability and safety during complex missions.

{\small
\bibliographystyle{IEEEtran}
\bibliography{references}

@article{asadi2020integrated,
  title={An integrated UGV-UAV system for construction site data collection},
  author={Asadi, Khashayar and Suresh, Akshay Kalkunte and Ender, Alper and Gotad, Siddhesh and Maniyar, Suraj and Anand, Smit and Noghabaei, Mojtaba and Han, Kevin and Lobaton, Edgar and Wu, Tianfu},
  journal={Automation in Construction},
  volume={112},
  year={2020},
  publisher={Elsevier}
}

@article{polka2017use,
  title={The use of UAV's for search and rescue operations},
  author={P{\'o}{\l}ka, Marzena and Ptak, Szymon and Kuziora, {\L}ukasz},
  journal={Procedia engineering},
  volume={192},
  pages={748--752},
  year={2017},
  publisher={Elsevier}
}

@article{hu2019detecting,
  title={Detecting, locating, and characterizing voids in disaster rubble for search and rescue},
  author={Hu, Da and Li, Shuai and Chen, Junjie and Kamat, Vineet R},
  journal={Advanced Engineering Informatics},
  volume={42},
  year={2019},
  publisher={Elsevier}
}

@book{roy2005rock,
  title={Rock blasting: effects and operations},
  author={Roy, Pijush Pal},
  year={2005},
  publisher={CRC Press}
}

@article{nedjati2016post,
  title={Post-earthquake response by small UAV helicopters},
  author={Nedjati, Arman and Vizvari, Bela and Izbirak, Gokhan},
  journal={Natural Hazards},
  volume={80},
  pages={1669--1688},
  year={2016},
  publisher={Springer}
}

@article{BAJPAYEE200447,
title = {Blasting injuries in surface mining with emphasis on flyrock and blast area security},
journal = {Journal of Safety Research},
volume = {35},
number = {1},
pages = {47-57},
year = {2004},
issn = {0022-4375},
doi = {https://doi.org/10.1016/j.jsr.2003.07.003},
url = {https://www.sciencedirect.com/science/article/pii/S0022437503000896},
author = {T.S. Bajpayee and T.R. Rehak and G.L. Mowrey and D.K. Ingram},
}

@techreport{mitchell2005risk,
  title={Risk assessment and update of inspection procedures for culverts},
  author={Mitchell, Gayle F and Masada, Teruhisa and Sargand, Shad M and Tarawneh, Bashar and others},
  year={2005},
  institution={United States. Federal Highway Administration}
}

@inproceedings{pushp2022uav,
  title={UAV-miniUGV Hybrid System for Hidden Area Exploration and Manipulation},
  author={Pushp, Durgakant and Kalhapure, Swapnil and Das, Kaushik and Liu, Lantao},
  booktitle={2022 IEEE/RSJ International Conference on Intelligent Robots and Systems (IROS)},
  pages={1297--1304},
  year={2022},
  organization={IEEE}
}

@misc{jaguar-mobile-platform,
  title = {Dr. Robot Jaguar Tracked Mobile Platform},
  howpublished = {\url{https://www.robotshop.com/products/dr-robot-jaguar-tracked-mobile-platform}},
  note = {Accessed: June 27, 2023}
}

@misc{packbot-robot,
  title = {PackBot},
  howpublished = {\url{https://www.wevolver.com/specs/packbot}},
  note = {Accessed: June 27, 2023}
}

@online{eyedrive,
    title = {EYEDRIVE - Tracked Robot Platform},
    url = {http://www.army-guide.com/eng/product5553.html},
    note = {Accessed: June 27, 2023}
}

@misc{dji_matrice600pro,
  author = {DJI},
  title = {Matrice 600 Pro - Product Specifications},
  howpublished = {\url{https://www.dji.com/matrice600-pro}},
  year = {Accessed 2023}
}

@inproceedings{pushp2023povnav,
  title={Povnav: A pareto-optimal mapless visual navigator},
  author={Pushp, Durgakant and Chen, Zheng and Luo, Chaomin and Gregory, Jason M and Liu, Lantao},
  booktitle={International Symposium on Experimental Robotics},
  pages={250--263},
  year={2023},
  organization={Springer}
}

@article{pushp2025navigating,
  title={Navigating the wild: Pareto-optimal visual decision-making in image space},
  author={Pushp, Durgakant and Chen, Weizhe and Chen, Zheng and Luo, Chaomin and Gregory, Jason M and Liu, Lantao},
  journal={The International Journal of Robotics Research},
  pages={02783649251399686},
  year={2025},
  publisher={SAGE Publications Sage UK: London, England}
}

@inproceedings{deng2020dbscan,
  title={DBSCAN clustering algorithm based on density},
  author={Deng, Dingsheng},
  booktitle={2020 7th international forum on electrical engineering and automation (IFEEA)},
  pages={949--953},
  year={2020},
  organization={IEEE}
}

@misc{qin2019a,
  Author = {Tong Qin and Jie Pan and Shaozu Cao and Shaojie Shen},
  Title = {A General Optimization-based Framework for Local Odometry Estimation with Multiple Sensors},
  Year = {2019},
  Eprint = {arXiv:1901.03638}
}

@ARTICLE{terrain_traversability,
  author={Borges, Paulo V. K. and Peynot, Thierry and Liang, Sisi and Arain, Bilal and Wildie, Matthew and Minareci, Melih G. and Lichman, Serge and Samvedi, Garima and Sa, Inkyu and Hudson, Nicolas and Milford, Michael and Moghadam, Peyman and Corke, Peter},
  journal={Field Robotics}, 
  title={A Survey on Terrain Traversability Analysis for Autonomous Ground Vehicles: Methods, Sensors, and Challenges}, 
  year={2022},
  volume={2},
  number={},
  pages={1567-1627},
  doi={10.55417/fr.2022049}}

@article{chen2022cali,
  title={Cali: Coarse-to-fine alignments based unsupervised domain adaptation of traversability prediction for deployable autonomous navigation},
  author={Chen, Zheng and Pushp, Durgakant and Liu, Lantao},
  journal={arXiv preprint arXiv:2204.09617},
  year={2022}
}

@article{chen2023pseudo,
  title={Pseudo-trilateral adversarial training for domain adaptive traversability prediction},
  author={Chen, Zheng and Pushp, Durgakant and Gregory, Jason M and Liu, Lantao},
  journal={Autonomous Robots},
  volume={47},
  number={8},
  pages={1155--1174},
  year={2023},
  publisher={Springer}
}

@article{khan2025afrda,
  title={AFRDA: Attentive Feature Refinement for Domain Adaptive Semantic Segmentation},
  author={Khan, Md Al-Masrur and Pushp, Durgakant and Liu, Lantao},
  journal={IEEE Robotics and Automation Letters},
  year={2025},
  publisher={IEEE}
}

@article{zou2018dynamic,
  title={Dynamic modeling and trajectory tracking control of unmanned tracked vehicles},
  author={Zou, Ting and Angeles, Jorge and Hassani, Ferri},
  journal={Robotics and Autonomous Systems},
  volume={110},
  pages={102--111},
  year={2018},
  publisher={Elsevier}
}

@inproceedings{pushp2024context,
  title={Context-Generative Default Policy for Bounded Rational Agent},
  author={Pushp, Durgakant and Xu, Junhong and Chen, Zheng and Liu, Lantao},
  booktitle={2024 IEEE/RSJ International Conference on Intelligent Robots and Systems (IROS)},
  pages={7703--7709},
  year={2024},
  organization={IEEE}
}

@inproceedings{xu2022decision,
  title={Decision-making among bounded rational agents},
  author={Xu, Junhong and Pushp, Durgakant and Yin, Kai and Liu, Lantao},
  booktitle={International Symposium on Distributed Autonomous Robotic Systems},
  pages={273--285},
  year={2022},
  organization={Springer}
}

@inproceedings{pushp2023coordination,
  title={Coordination of bounded rational drones through informed prior policy},
  author={Pushp, Durgakant and Xu, Junhong and Liu, Lantao},
  booktitle={2023 IEEE/RSJ International Conference on Intelligent Robots and Systems (IROS)},
  pages={6779--6786},
  year={2023},
  organization={IEEE}
}
}

\end{document}